\documentclass[10pt,journal,compsoc]{IEEEtran}
\usepackage{tabularx}
\usepackage{caption}
\usepackage{subcaption}
\usepackage{pifont}
\usepackage{graphicx}
\usepackage{pgfplots}
\usepackage{hyperref}
\pgfplotsset{compat=1.16}
\usepgfplotslibrary{statistics}
\usepackage{fancyhdr}
\usepackage{amsmath, xparse}
\usepackage{xcolor,colortbl}
\usepackage{booktabs}

\ifCLASSOPTIONcompsoc
  \usepackage[nocompress]{cite}
\else
  \usepackage{cite}
\fi
\ifCLASSINFOpdf
\else
\fi
\begin{document}
%
\title{Towards Robust and Unconstrained Full Range of Rotation Head Pose Estimation}
%
%
%
%

\author{Thorsten Hempel,
        Ahmed A. Abdelrahman
        and Ayoub Al-Hamadi
\IEEEcompsocitemizethanks{\IEEEcompsocthanksitem T. Hempel, A. A. Abdelrahman and A. Al-Hamadi are with the Faculty of Electrical Engineering and Information Technology, Otto-von-Guericke University, Magdeburg, 39106, Germany.\protect\\
}
\thanks{Manuscript received April 19, 2005; revised August 26, 2015.}}

%
%


\markboth{\shortstack{This work has been submitted to the IEEE for possible publication. Copyright may be transferred without notice,  \\
after which this version may no longer be accessible.
}}{}
%



\def\x{{\mathbf x}}
\def\L{{\cal L}}
\newcommand{\cmark}{\ding{51}}%
\newcommand{\xmark}{\ding{55}}%
\newcommand\netname{6DRepNet}
\IEEEtitleabstractindextext{%
\begin{abstract}
Estimating the head pose of a person is a crucial problem for numerous applications that is yet mainly addressed as a subtask of frontal pose prediction. We present a novel method for unconstrained end-to-end head pose estimation to tackle the challenging task of full range of orientation head pose prediction. We address the issue of ambiguous rotation labels by introducing the rotation matrix formalism for our ground truth data and propose a continuous 6D rotation matrix representation for efficient and robust direct regression. This allows to efficiently learn full rotation appearance and to overcome the limitations of the current state-of-the-art. Together with new accumulated training data that provides full head pose rotation data and a geodesic loss approach for stable learning, we design an advanced model that is able to predict an extended range of head orientations. An extensive evaluation on public datasets demonstrates that our method significantly outperforms other state-of-the-art methods in an efficient and robust manner, while its advanced prediction range allows the expansion of the application area.
We open-source our training and testing code along with our trained models: \url{https://github.com/thohemp/6DRepNet360}.
\end{abstract} 

}

\maketitle

\IEEEdisplaynontitleabstractindextext

%
\IEEEpeerreviewmaketitle

\IEEEraisesectionheading{\section{Introduction}\label{sec:introduction}}

%
%
%
%
\IEEEPARstart{H}{head} pose estimation follows the objective of predicting the human head orientation from images and is a crucial step in many computer vision algorithms. Applications are wide-ranging and include attention estimation~\cite{Veronese2017ProbabilisticMO, visfocus, 9570335}, face recognition~\cite{Chang2017FacePoseNetMA, facerecog}, and the estimation of facial attributes~\cite{KUMAR201849,Ranjan2019HyperFaceAD}, which again are vital features in driver assistance systems~\cite{4357803, driverhp, poseidon}, augmented reality~\cite{5443483, buehler2021varitex}, and human-robot interaction~\cite{app11125366, Gaschler2012ModellingSO, engagement}. The vast majority of present methods~\cite{Ruiz2018FineGrainedHP, Yang_2019_CVPR, Huang2020ImprovingHP, 8444061, Zhang2020FDNFD,9435939,10.1007/978-3-030-90439-5_43,rankpose} narrow down the research issue to the estimation of solely frontal poses with a limited rotation range. This favors the leverage of the facial feature-richness and suitable, widely available training datasets. However, in uncontrolled application scenarios~\cite{8448756,10.1109/ICIP.2015.7351449,7254213} head orientations are likely to surpass the narrow angle range that most methods are trained for and, consequently, produce random and inaccurate head pose predictions. In view of extending the prediction to the full area of rotation range, the current state of research is challenged by two key limitations. The first is the absence of comprehensive datasets that cover the full range of head orientations~\cite{KHAN2021116479}. The second equally decisive and often neglected factor is an appropriate rotation representation, as it significantly impacts the model's ability to effectively learn the connection between visual pose appearance and corresponding parameterization~\cite{Zhou2019OnTC}. For instance, the commonly used Euler angle and quaternion representation suffer from ambiguity and discontinuity problems that lead to an unstable training process and a mediocre prediction performance if plainly applied~\cite{Ruiz2018FineGrainedHP, 8444061, Zhou2020WHENetRF, rankpose}. This behavior even intensifies for stronger rotations in the narrow range spectrum. 
\begin{figure}[t]
\centering
  \begin{subfigure}{0.24\linewidth}
    \includegraphics[width=\linewidth]{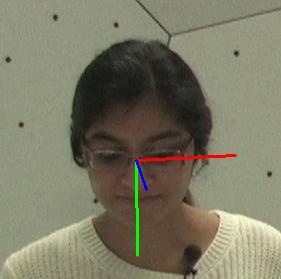}
  \end{subfigure}
  \begin{subfigure}{0.24\linewidth}
    \includegraphics[width=\linewidth]{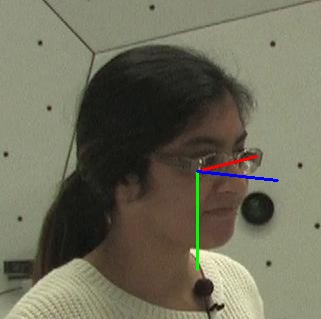}
  \end{subfigure}
  \begin{subfigure}{0.24\linewidth}
    \includegraphics[width=\linewidth]{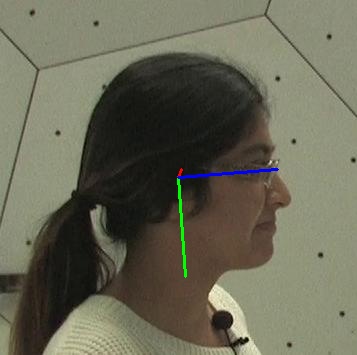}
  \end{subfigure}
  \begin{subfigure}{0.24\linewidth}
    \includegraphics[width=\linewidth]{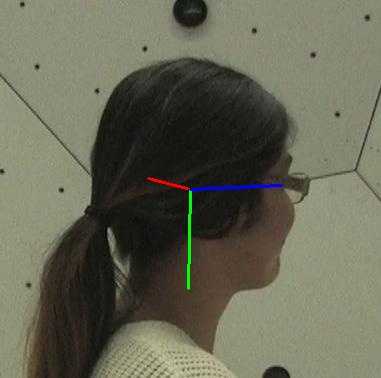}
  \end{subfigure}
    \begin{subfigure}{0.24\linewidth}
    \includegraphics[width=\linewidth]{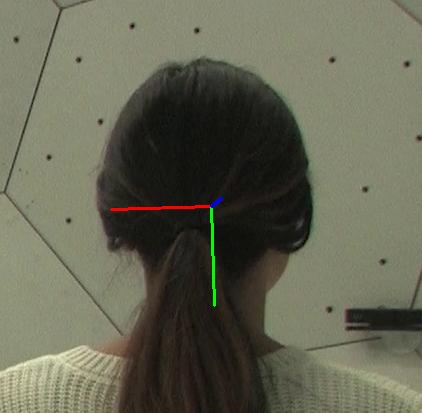}
  \end{subfigure}
    \begin{subfigure}{0.24\linewidth}
    \includegraphics[width=\linewidth]{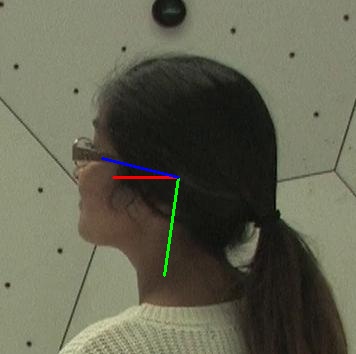}
  \end{subfigure}
   \begin{subfigure}{0.24\linewidth}
    \includegraphics[width=\linewidth]{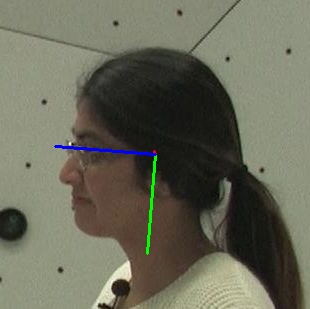}
  \end{subfigure}
    \begin{subfigure}{0.24\linewidth}
    \includegraphics[width=\linewidth]{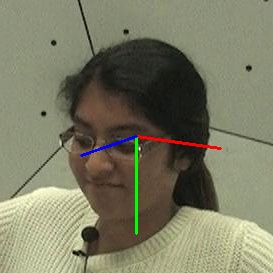}
  \end{subfigure}
  \caption{Example images of predicted orientations of various rotated heads.}
  \label{fig:intro_results}
\end{figure}

We overcome these limitations by proposing a rotation matrix-based 6D representation for efficient and unconstrained network training that we further enhance with a geodesic based loss. 
Additionally, we take up the ambitious challenge of predicting the full range of rotation by agglomerating new training data with enhanced pose variation. For this matter, we utilize the CMU Panoptic~\cite{jooiccv2015} dataset and apply an automatic head pose labeling process to generate head pose samples with focus on the back of the head. We combine these samples with the popular 300W-LP~\cite{Zhu2016FaceAA} head pose dataset and, together, receive a large scaled dataset with greatly expanded head rotation variations. Finally, the training of our proposed model on this new agglomerated data enables us to predict a significantly extended range of head orientations. We examine our approach in multiple experiments on public datasets that testify our method state-of-the-art accuracy and remarkable robustness in predicting challenging poses. At the same time, it is able to handle a many times greater range of head pose orientations compared to current methods from the literature. Fig~\ref{fig:intro_results} shows examples of orientation predictions from this model for versatile head poses. To the best of our knowledge, we are the first to tackle the full range of head pose estimation in this extensive and conclusive way. In summary, we make the following contributions:

\begin{itemize}
    \item We introduce a simplified and efficient 6-parameter rotation matrix representation for regressing accurate head orientations without suffering ambiguity problems.
    \item We propose a geodesic distance approach for network penalizing to encapsulate the training loss within the Special Orthogonal Group \textit{SO}(3) manifold geometry. 
    \item We utilize the CMU Panoptic dataset~\cite{jooiccv2015} to expand the traditional 300W-LP~\cite{Zhu2016FaceAA} head pose dataset with full rotation head pose appearance.
    \item We create a new head pose prediction model that surpasses the prediction range of current methods and at the same time achieves lower errors on common test datasets. 
    \item We demonstrate the superiority of our approach in accuracy and robustness in multiple experimental setups.
    \item We conduct an ablation study to evaluate the impact of each component of our model on the achieved results.
\end{itemize}

\begin{figure*}
    \centering
    \includegraphics[width=\linewidth]{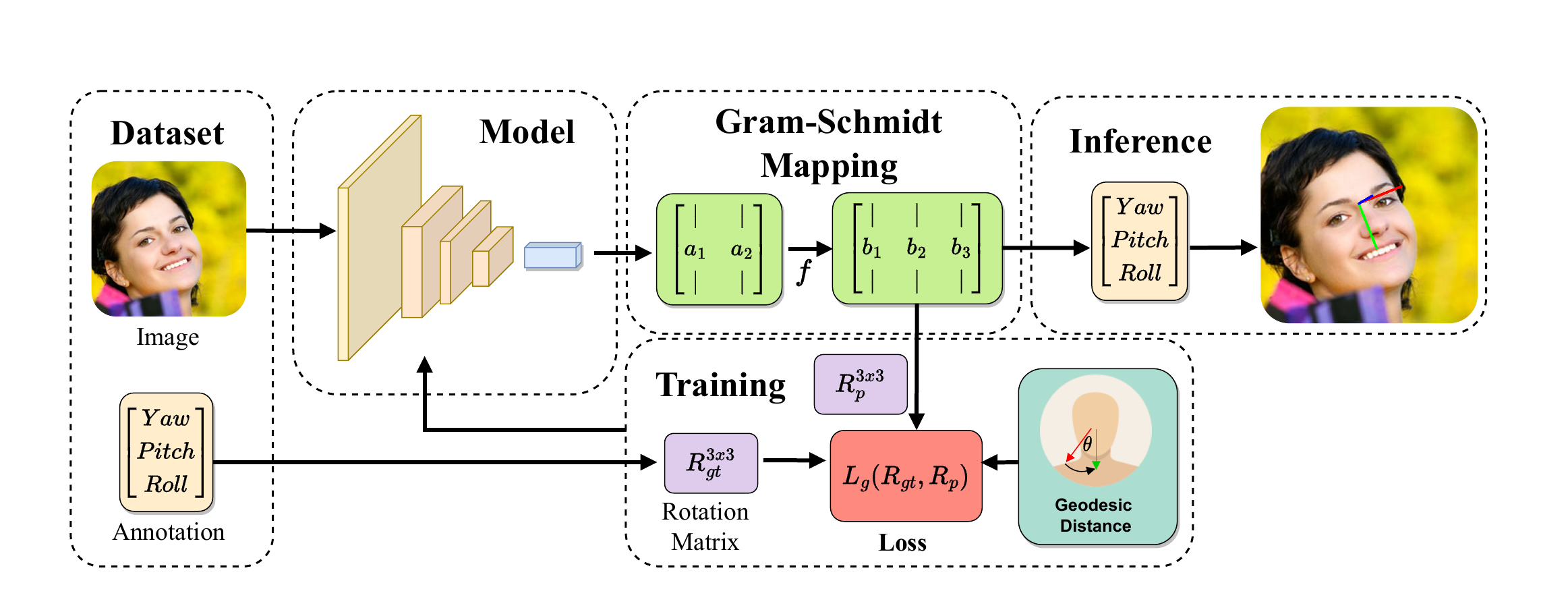}
    \caption{Overview of the proposed method.}
    \label{fig:method}
\end{figure*}

Fig \ref{fig:method} shows an overview of our proposed method. Each component will be explained in detail in the following sections. Inspired by the 6D representation that is used in our approach, we call our network \netname. An earlier version of this work was published in \cite{https://doi.org/10.48550/arxiv.2202.12555}(\textit{accepted to ICIP2022}), where we presented an initial approach for 6D-based narrow angle prediction. In this version, we enhance this previous work with an improved training procedure, propose an approach for tackling the prediction of the full range of orientation, and provide a more detailed model including an extensive comparison with the state-of-the-art, error analysis and ablation studies.

Our training, testing code, and trained models are made publicly available to facilitate research experimentation and practical application development.
\section{Related Work}
\label{rel_work}
In recent years, facial analysis along with vision-based head orientation prediction emerged with the rise of neural networks. Current methods are commonly divided into landmark-based and landmark-free approaches. Landmark-based methods~\cite{bulat2017far,8297015, 6909637, neehpe} detect facial landmarks as a primary step and subsequently recover the 3D head pose by aligning the predicted landmarks with a standardized 3D head model~\cite{Li2012ARO,1699072}. Under ideal circumstances, this approach can lead to very accurate head orientation estimations, but it is highly dependent on the precise predictions of the landmark positions. Also, it requires the target head to be shaped similar to the head model to achieve an accurate alignment. Finally, as the target landmarks are only located in the facial area, poses with occlusion and of strong rotated heads with too few or without visible face cannot be estimated~\cite{6751298,7410774}.
Landmark-free approaches overcome these limitations by directly estimating the head pose from the images in an end-to-end fashion. These methods commonly use deep neural networks to formulate the orientation prediction as an appearance-based task. As one of the first of its kind, HopeNet~\cite{Ruiz2018FineGrainedHP} presented an RvC~\cite{10.1007/3-540-61859-7_6} approach by binning the target angle range to combine a cross-entropy and a mean squared error loss function for Euler angle prediction. Along with this classification approach, they at the same time reduced the predictable rotation range between [-99,+99] degree for yaw, pitch, and roll. Later, QuatNet~\cite{8444061} adapted the cross-entropy paradigm with limited prediction range and proposed to split classification and regression into separate network branches. One branch is used for classifying the Euler angles and the second one regresses the pose in quaternion representation. Similarly, HPE~\cite{Huang2020ImprovingHP} treats classification and regression separately and averages the outputs as a pose regression subtask. WHENet~\cite{Zhou2020WHENetRF} keeps the single branch strategy, switches to an EfficientNet~\cite{pmlr-v97-tan19a} backbone and increases the number of bins for the yaw network branch to extend the predictable angle range. Whereas FSA-Net~\cite{Yang_2019_CVPR} proposes a network with a stage-wise regression and feature aggregation scheme for predicting Euler angles. TriNet~\cite{Cao_2021_WACV} adapts this method, but estimates the three unit vectors of the rotation matrix instead of Euler angles and incorporates an additional orthogonality loss to stabilize the predictions. MFDNet~\cite{9435939} likewise follows the rotation matrix representation but uses its Fisher distribution to model rotation uncertainty and to find its maximum likelihood. Another probabilistic approach was proposed by Liu et al.~\cite{9022536} who train on Gaussian label distributions. Whereas FDN~\cite{Zhang2020FDNFD} targets optimized feature extraction by proposing a feature decoupling method to explicitly learn discriminative features of different head orientations. DDD-Pose~\cite{10.1007/978-3-030-90439-5_43} seeks to diversify the training data by proposing an advanced augmentation scheme. The current state-of-the-art results are achieved by RankPose~\cite{rankpose} closely followed by MNN~\cite{Valle2021MultiTaskHP}. RankPose uses paired training samples to introduce a ranking loss for penalizing incorrect ordering of the Euler pose estimation. MNN and img2pose~\cite{9578214} predict the rigid transformation between the head and the camera.

In general, frequent approaches in the area of head pose estimation achieved continuous improvement over the recent years, yet they still lack of comprehensive solutions for predicting the full range of head pose rotation. First, it became a common convention to split up the continuous rotation variables into bins to convert the problem into a classification task in order to stabilize the predictions~\cite{Ruiz2018FineGrainedHP,8444061,Huang2020ImprovingHP,Zhou2020WHENetRF,Zhang2020FDNFD}. However, this is problematic as pruning segments of angles into bins will consequently lead to a loss of information. Apart from that, this constraining approach is commonly combined with reducing the target space~\cite{Ruiz2018FineGrainedHP,8444061,Huang2020ImprovingHP,Zhou2020WHENetRF,Zhang2020FDNFD} which eliminates the opportunity of tackling full rotation estimation. A few works overcome these limiting factors by using the rotation matrix as a more suitable rotation representation~\cite{Cao_2021_WACV, 9435939,Viet2021SimultaneousFD}, but neither deal with more efficient ways of regression nor address its potential for expanding the prediction range. As a consequence, the area of full head pose prediction is still rarely explored yet. WHENet~\cite{Zhou2020WHENetRF} was one of the first to approach full yaw prediction by extending the bin range for the yaw angle and proposing a wrapping loss to handle the influence of the gimbal lock. However, their method still tightly restricts pitch and roll between [-99,+99] degrees. The same restriction is applied by Viet et al.~\cite{Viet2021SimultaneousFD} in their multitask approach, where they face detection and head pose estimation. As rotation representation, they use the rotation matrix and follow the same computational extensive approach as TriNet~\cite{Cao_2021_WACV} to obtain orthogonality.

\section{Method}
In the following, we will give details about our proposed method. We start with preliminary information about different rotation representations. Based on its insights, we propose a rotation parametrization scheme to overcome the limitation of the related works. As an accompanying measure, we will introduce a geodesic distanced based loss to precise and stabilize the network penalty for training.
\begin{figure}[]
\scalebox{0.8}{
\begin{tabular}{cc}
\includegraphics[width=.45\linewidth]{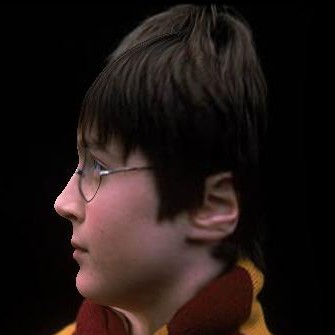} &
\includegraphics[width=.45\linewidth]{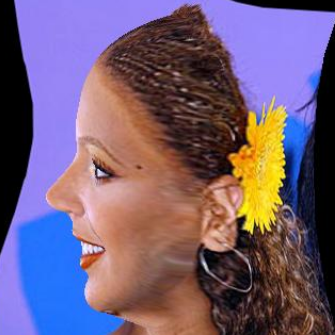}\\
\addlinespace[2ex]
\textbf{Euler Angles} & \textbf{Euler Angles} \\
 \addlinespace[1ex]
$\begin{bmatrix} 87.73 &  89.32   &-87.93 \end{bmatrix}$ &
$\begin{bmatrix}-92.51 &  -98.02  & 81.73\end{bmatrix}$   \\
\addlinespace[2ex]
\textbf{Quaternions} & \textbf{Quaternions}\\
\addlinespace[1ex]
$\begin{bmatrix}
0.707  & -0.023  & -0.707  & 0.031
\end{bmatrix}                                                                                     
$
& 
$\begin{bmatrix}-0.707 &  -0.029   &0.706  & 0.041\end{bmatrix}$  \\
 \addlinespace[2ex]
\textbf{Rotation matrix} & \textbf{Rotation matrix} \\
\addlinespace[1ex]
$\begin{bmatrix} \mathrm{0.000} &  0.012 &  -0.999\\  -0.076 &  -0.997  & -0.012\\  -0.997  &  0.076 &  0.000 \end{bmatrix} $ & 
$\begin{bmatrix} 0.002  & -0.017 &  -0.999\\ 0.100 &  -0.995 &  0.017\\  -0.995  & -0.100 &  -0.001 \end{bmatrix} $
\end{tabular}
}
\caption{Data samples from 300W-LP dataset with different rotation parameterization.}
\label{fig:rot_parameterization}
\end{figure}
\subsection{Preliminaries}
\label{sec:rot_rep}
In general, the orientation of a rigid body in the three-dimensional space can be described by multiple kinds of mathematical representation. The most common and widely used one is the Euler angle representation that is used to describe the rotation around each axis of the coordinate system (typical denoted as $yaw, pitch, roll$). Despite its intuitiveness, Euler angles face limitations when it comes to the specific orientation state, where the second elemental rotation reaches 90 or -90 degrees. Given this setup, $yaw$ and $pitch$ align on the same plane and create infinitive solutions for the same rotation state. This behavior is known as $gimbal~lock$ as the first and third axis are locked under this particular condition. The gimbal lock represents the extreme case for the limitations of Euler angles. However, the dependency between first and third angle is a fundamental property of Euler angles, that just becomes stronger the more the pitch reaches the gimbal lock state. As a consequence, the Euler angle representation does not behave in the same continuous form as its visual appearance counterpart that has a detrimental impact on the performance of neural networks.

Another type of orientation is called axis-angle representation, which consists of a unit vector $v=(\tilde{x}, \tilde{y}, \tilde{z})$ that defines the axis of the rotation and an angle $\theta$ that describes the magnitude of its rotation. Closely related to the axis-angle representation, another type called rotation quaternions $q$ with also four parameters $q_0, q_1, q_2, q_3$ can be derived by $q_0=cos(\frac{\theta}{2}), q_1=\tilde{x}~sin(\frac{\theta}{2}),q_2=\tilde{y}~sin(\frac{\theta}{2}),q_3=\tilde{z}~sin(\frac{\theta}{2})$. Quaternions and the axis-angle representation are not affected by the gimbal lock, but they still have an ambiguity that is introduced by their antipodal symmetry with $-v=v$ and $-q = q$, respectively. As a result, every orientation can be described by two different representations that are maximum far apart.   
A more comprehensive notation is the rotation matrix $R^{3x3}$ that consists of 9 parameters. Despite its increased number of parameters, it comes with the crucial advantage that it provides a continuous representation with a unique parameterization for each rotation. Fig.~\ref{fig:rot_parameterization} shows an example of two dataset samples with similar pose appearances. Yet, their Euler angle and quaternion ground truth are parameterized very differently. Only the rotation matrices reflects the similarity in the pose appearance.
In \textit{SO}(3) the matrix representation $R$ is sized $3\times3$ with an orthogonality constraint $RR^T=I$, where $R^T$ is the transposed matrix and $I$ the identity matrix.
\begin{equation}
    R = \begin{bmatrix}
    r_{11} & r_{12} & r_{13} \\
    r_{21} & r_{22} & r_{23} \\
    r_{31} & r_{32} & r_{33} \\
\end{bmatrix}
 \end{equation}
One could now try to regress the rotation matrix directly, but this would require finding all nine parameters that at the same time satisfy the orthogonality constraint. The orthogonality can also be enforced in a sequential step by either using the Gram-Schmidt process or the singular value decomposition (SVD). The SVD is an extensive approach for finding those orthogonal vectors that are the nearest to the predictions. The Gram-Schmidt method requires discarding one vector in order to recreate the orthogonal matrix from the remaining two. 
\subsection{6D Representation}
In section~\ref{sec:rot_rep} we show that a key aspect for tackling direct orientation predictions is the use of an appropriate rotation representation that is unambiguously interpretable by neural networks. For this matter, we use the rotation matrix representation as a superior alternative to Euler angles, quaternions, and axis-angles. 
Inspired by Zhou et al.~\cite{Zhou2019OnTC}, we satisfy the orthogonality constraint by performing the Gram-Schmidt mapping inside the representation itself, which avoids extensive post-processing.
We simply drop the last column vector of the rotation matrix that reduces the $3\times3$ matrix into a 6D rotation representation
\begin{equation}
\label{eq:gs_mapping}
    g_{GS}=\left( \left[ \begin{matrix}| & | & | \\ a_1 & a_2 & a_3 \\ | & | & | \\\end{matrix} \right]\right) = \left[  \begin{matrix}   | & | \\ a_1 & a_2  \\ | & |   \end{matrix}\right],
\end{equation}
which has been reported to introduce smaller errors for direct regression~\cite{Zhou2019OnTC}. Then, the predicted 6D representation matrix is mapped back into \textit{SO}(3) with
\begin{equation}
    f_{GS}=\left( \left[\begin{matrix}| & |  \\ a_1 & a_2  \\ | & |  \\\end{matrix}  \right]\right) = \left[ \begin{matrix}| & | & | \\ b_1 & b_2 & b_3 \\ | & | & | \\\end{matrix}\right],
\end{equation}
where the resulting column vectors are defined as
\begin{equation}
\begin{split}
& b_1 = \frac{a_1}{||a_1||}, \\
& b_2 = \frac{u_2}{||u_2||} \textit{ with } u_2=a_2-(b_1\cdot a_2)b_1,\\
& b_3 = b_1\times b_2.\\
\end{split}
\end{equation}
Hereby, the last column vector is simply determined by the cross product that ensures that the orthogonality constraint is satisfied for the resulting $3\times3$ matrix:

As a result, our network has only to predict 6 parameters that are mapped into a $3\times3$ rotation matrix in a subsequent transformation process that incorporates the orthogonality constraint as well.

\subsection{Geodesic loss}
The \textit{l2}-norm is the commonly used loss function for head pose related tasks. However, using the Frobenius norm for measuring distances between two matrices would break with the \textit{SO}(3) manifold geometry. Instead, the shortest path between two 3D rotations is geometrically interpreted as the geodesic distance. Let $R_p$ and $R_{gt}$ $\in$ \textit{SO}(3) be the estimated and the ground truth rotation matrices, respectively, then the geodesic distance between both rotation matrices is defined as:

\begin{equation}
    L_{g}=cos^{-1}\left( \frac{tr(R_p R_{gt}^T)-1}{2}\right).
    \label{label:geodesicloss}
\end{equation}
In the following, we will use this metric as a loss function for our neural network to compute accurate distance information between the predicted and ground truth orientation.

\section{Experiments}
\label{sec:experiments}
\begin{figure}
    \centering
    \begin{subfigure}{.8\linewidth}
        \includegraphics[width=\linewidth]{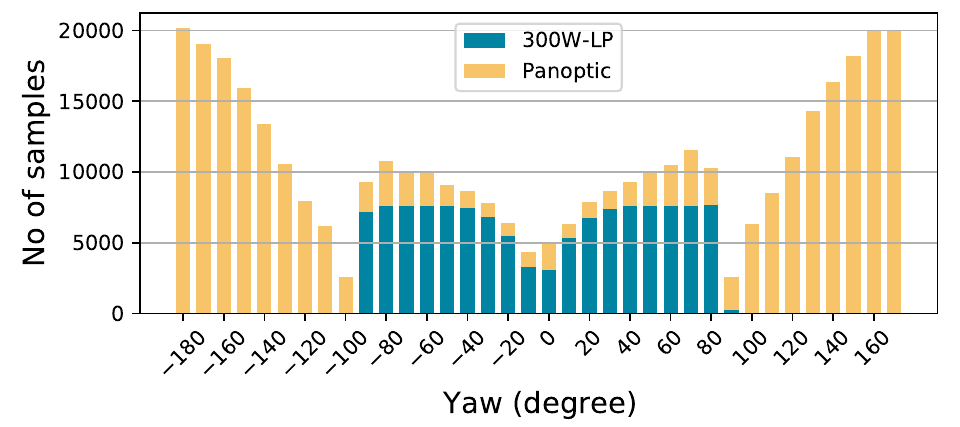}
    \end{subfigure}
    \begin{subfigure}{.8\linewidth}
        \includegraphics[width=\linewidth]{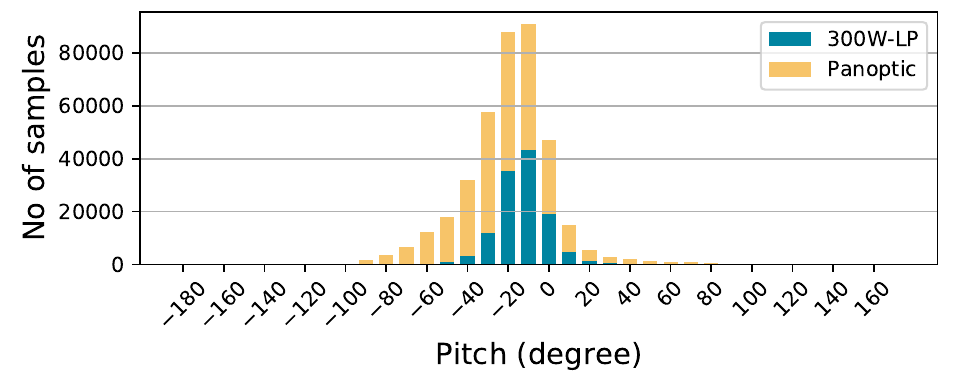}
    \end{subfigure}
    \begin{subfigure}{.8\linewidth}
        \includegraphics[width=\linewidth]{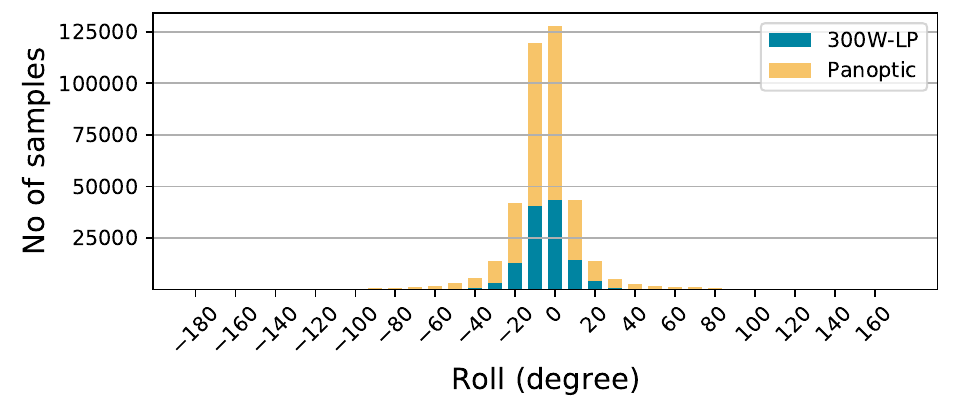}
    \end{subfigure}
    \caption{Dataset distribution}
    \label{fig:dataset_distribution}
\end{figure}


\begin{figure*}[t]
\centering
  \begin{subfigure}{0.16\linewidth}
    \includegraphics[width=\linewidth]{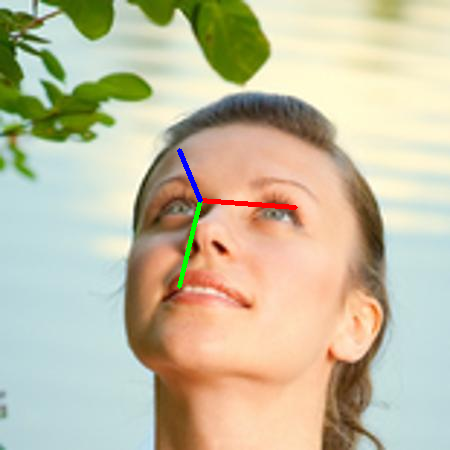}
  \end{subfigure}
  \begin{subfigure}{0.16\linewidth}
    \includegraphics[width=\linewidth]{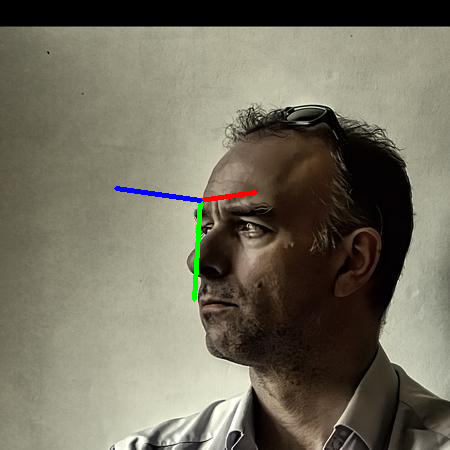}
  \end{subfigure}
  \begin{subfigure}{0.16\linewidth}
    \includegraphics[width=\linewidth]{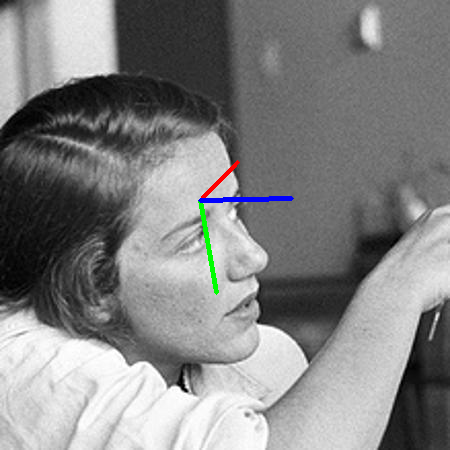}
  \end{subfigure}
  \begin{subfigure}{0.16\linewidth}
    \includegraphics[width=\linewidth]{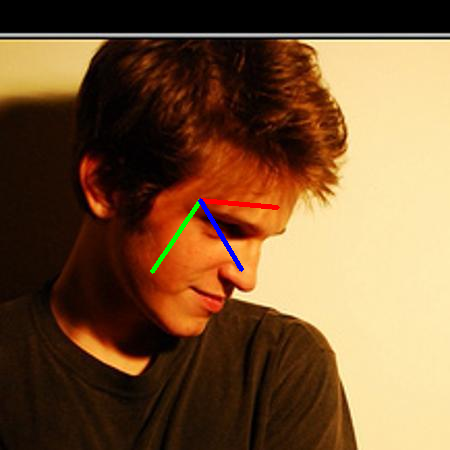}
  \end{subfigure}
    \begin{subfigure}{0.16\linewidth}
    \includegraphics[width=\linewidth]{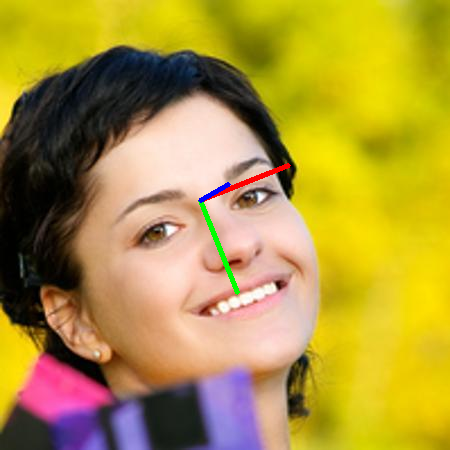}
  \end{subfigure}
    \begin{subfigure}{0.16\linewidth}
    \includegraphics[width=\linewidth]{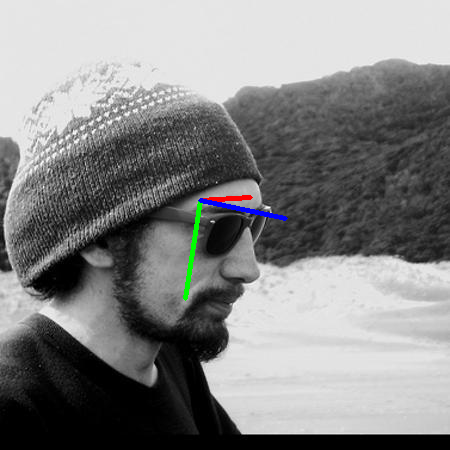}
  \end{subfigure}
  
  \vspace{0.2em}
  
    \begin{subfigure}{0.16\linewidth}
    \includegraphics[width=\linewidth]{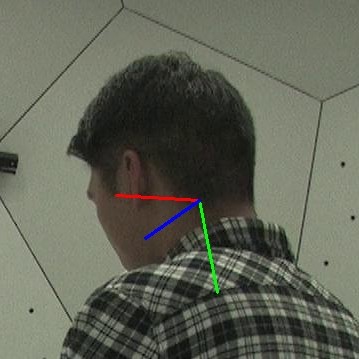}
  \end{subfigure}
  \begin{subfigure}{0.16\linewidth}
    \includegraphics[width=\linewidth]{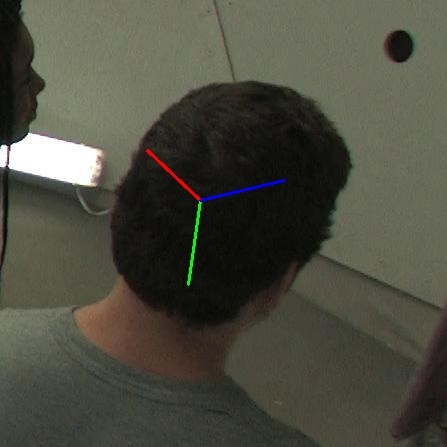}
  \end{subfigure}
  \begin{subfigure}{0.16\linewidth}
    \includegraphics[width=\linewidth]{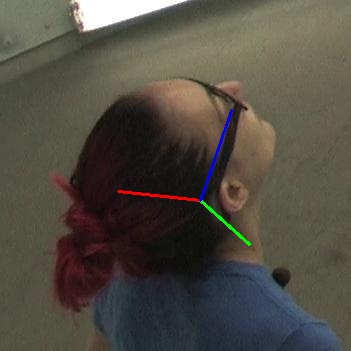}
  \end{subfigure}
  \begin{subfigure}{0.16\linewidth}
    \includegraphics[width=\linewidth]{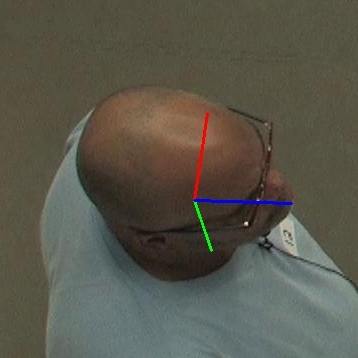}
  \end{subfigure}
    \begin{subfigure}{0.16\linewidth}
    \includegraphics[width=\linewidth]{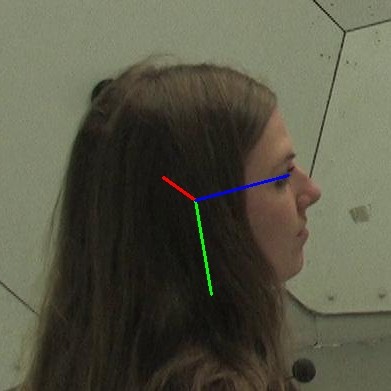}
  \end{subfigure}
    \begin{subfigure}{0.16\linewidth}
    \includegraphics[width=\linewidth]{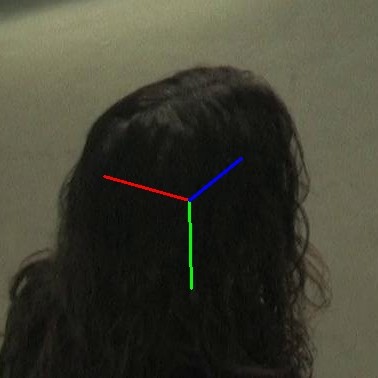}
  \end{subfigure}
  \caption{Example images with converted Euler angle visualization from the AFLW2000 dataset (first row) and the CMU Panopic test dataset (second row).}
  \label{fig:qual_results}
\end{figure*}

We perform an extensive evaluation of our method. We begin the specification of our used datasets, evaluation metrics and implementation setup, followed by a comprehensive comparison with other state-of-the-art methods in cross-dataset and intra-dataset tests. Further analysis includes a detailed error analysis and ablation studies on used loss functions and backbones.
\subsection{Datasets}
We conduct our evaluation with the aid of different kinds of data. The most common and public available datasets are 300W-LP~\cite{Zhu2016FaceAA}, AFLW2000~\cite{7298679}, and BIWI~\cite{FDGF12}.

\textbf{300W-LP}: 300W-LP consists of 66,225 face samples collected from multiple databases including LFPW~\cite{fiducials_pami2013}, AFW~\cite{Bulat2016TwoStageCP}, HELEN~\cite{Le2012InteractiveFF} and iBUG~\cite{6595977} that are further enhanced to 122,450 samples by image flipping. It is based on around 4000 real images. The ground truth is provided in the Euler angle format. For training, we convert them into the matrix form.

\textbf{AFLW2000}: The AFLW2000 dataset contains the first 2,000 images from the ALFW dataset annotated with the ground truth 3D faces and the corresponding 68 landmarks. It contains samples with large variations, different illumination, and occlusion conditions.

\textbf{BIWI}: The BIWI dataset includes 15,678 images that were created in a lab environment with 20 participants. In this dataset, the head takes up only a small area in the images. Hence, we use the MTCNN~\cite{Zhang2016JointFD} face detector to loosely crop the heads from the images.
All of the above listed datasets provide, due to their nature of annotation, only samples with a frontal view of the faces (mostly between -99° and +99° range of yaw). Therefore, they cannot be used for the training of the entire head orientation range. 

\textbf{CMU Panoptic}: Therefore, we utilize another dataset called CMU Panoptic~\cite{jooiccv2015} that makes it possible to generate annotated head images with full rotation appearances. 
In this dataset, a variety of subjects perform arbitrary tasks inside a dome, that is equipped with 31 evenly arranged HD cameras. The main focus of this dataset is to capture the subject's poses, but it also provides 3D facial landmark annotation and camera intrinsicts and extrinsics. This enables to extract head pose annotation from all the different camera angles, that was initially harnessed by Zhou et al.~\cite{Zhou2020WHENetRF}.
There are 30 sequences public available with multiple subjects per scene, that are standing in a ring with each subject being oriented towards the center of the dome. When extracting the head crops with only accepted those with a minimum size of 320 for both axis, which gives us a dataset with 113914 samples in total. Because of the subject's spacial setup, the majority of the samples are ones showing the back of the head. Samples with frontal face view are likelier to be sorted out by too small sized, as these face images were taken from more far distance. Therefore, we create a combination of the 300W-LP and the CMU Panoptic dataset that includes 236,364 data samples spanning the entire range of yaw rotation. The range of pitch is slightly expanded as we also use the samples that are generated from cameras attached to the ceiling of the CMU Panoptic dome. The distribution of this new training data is shown in Fig. \ref{fig:dataset_distribution}. It should be noted that we use the Euler angles for presentation purposes that cannot exactly represent the distribution of visual appearance in the dataset, as discussed in section \ref{sec:rot_rep}.

\subsection{Evaluation Metrics}
We use two different evaluation metrics to quantify the head pose estimations error. The first one is the most common Mean Absolute Error (MAE) of the Euler angles,
\begin{equation}
    \text{MAE} = \frac{1}{N} \sum_{i=1}^{N}(|x_g - x_p|),
\end{equation}
where $N$ is the number of face images and $x_g$ and $x_p$ represent the ground truth and predicted pose parameters, respectively.
Secondly, we calculate the Mean Absolute Error of the vectors (MAEV) of the rotation matrix. This metric was introduced by \cite{Cao_2021_WACV} in order to surpass the limitations of the Euler representation and to provide a more meaningful picture of the appearance differences between predicted and ground truth orientation. The MAEV defines the angle error between the three vectors of the rotation matrix, 
\begin{equation}
    \text{MAEV} = \frac{1}{N} \sum_{i=1}^{N}cos^{-1} \left ( \frac{v_g \cdot v_p}{|v_g||v_p|} \right ),
\end{equation}
where $N$, again, is the number of face images in the dataset and $v_g$ and $v_p$ are the ground truth and the predicted head orientation vectors.
\definecolor{Gray}{gray}{0.85}
\newcolumntype{a}{>{\columncolor{Gray}}c}
\begin{table*}[h!]
 \setlength{\tabcolsep}{3.9pt}
\centering
\begin{tabular} { l c c c c c a c c c a   c c  c a   c c c a  }
\toprule
& & & \multicolumn{8}{c}{\textbf{Euler}} & \multicolumn{8}{c}{\textbf{Vector}} \\
\midrule
& & & \multicolumn{4}{c}{\textbf{AFLW2000}} & \multicolumn{4}{c}{\textbf{BIWI}} & \multicolumn{4}{c}{\textbf{AFLW2000}} & \multicolumn{4}{c}{\textbf{BIWI}}\\
 \midrule
& I & R &  Yaw & Pitch & Roll & MAE & Yaw & Pitch & Roll & MAE  & Left & Down & Front & MAEV & Left & Down & Front & MAEV  \\
 \midrule
 HopeNet~\cite{Ruiz2018FineGrainedHP}& \cmark & \xmark & 6.40 & 6.53 & 5.39 &  6.11 &      4.54 & 5.15 & 3.37 & 4.36 &     7.98 & 6.40 & 8.54 & 7.64 &     6.21 & 5.75 & 7.05 & 6.33\\ 
 FSA-Net~\cite{Yang_2019_CVPR}& \cmark & \xmark &  4.83 & 6.25 & 4.94 & 5.34 & 4.64 & 5.61 & 3.57 & 4.61 & 6.88 & 6.52 & 7.28 & 6.89 & 6.27 & 6.29 & 7.38 & 6.65 \\ 
 HPE~\cite{Huang2020ImprovingHP}& \xmark&  \xmark  &  4.80 & 6.18 & 4.87 & 5.28 & 3.12 & 5.18 & 4.57 & 4.29 &     - & -& -& -&    -& -&-&-\\
 QuatNet~\cite{8444061}& \xmark& \xmark & 3.97 & 5.62  & 3.92 & 4.50 & \textbf{2.94} & 5.49 & 4.01 & 4.15 &    -& -&-&-&   -&-&-&- \\
 TriNet~\cite{Cao_2021_WACV}& \cmark& \xmark & 4.36 & 5.81 & 4.51 & 4.89 & 3.11 & 5.09 & 5.20 & 4.47 &  6.16 & 5.95 & 6.82 & 6.31 &  6.58 & 5.80 & 7.55 & 6.64 \\
 WHENet-V~\cite{Zhou2020WHENetRF}& \xmark & \xmark &4.44& 5.75& 4.31& 4.83 & 3.60 & 4.10 & 2.73 & 3.48&   - & -& -& -&    -& -&-&-\\
 WHENet~\cite{Zhou2020WHENetRF}& \xmark & \cmark &  5.11 & 6.24 & 4.92 & 5.42 & 3.99 & 4.39 & 3.06 & 3.81&     - & -& -& -&    -& -&-&-\\
 FDN~\cite{Zhang2020FDNFD} & \xmark& \xmark  & 3.78 & 5.61 & 3.88 & 4.42 & 4.52 & 4.70 & \textbf{2.56} & 3.93    & -&-&-&-&      -&-&-&-\\
 Viet et al~\cite{Viet2021SimultaneousFD}& \xmark& \cmark  &  -& -& -& -& 4.62 &4.29 &4.52 &4.48 & -&-&-&-&-&-&-&-\\
 MFDNet~\cite{9435939}& \xmark& \xmark  &  4.30 & 5.16 & 3.69 & 4.38 & 3.40 & 4.68 & 2.77 & 3.62 & -&-&-&-&-&-&-&- \\
 DDD-Pose~\cite{10.1007/978-3-030-90439-5_43}& \xmark & \xmark &4.38 & 4.85 & 3.44 & 4.22 & 4.60 & 6.02 & 2.94 & 4.52 &- &-&-&-&-&-&-&- \\
 Liu et al.~\cite{9022536}& \xmark & \xmark &  3.03 & 5.06 & 3.68 & 3.93 & 4.12 & 5.61 & 3.15 & 4.29 & -& -& -& -& -& -& -& - \\
 img2pose~\cite{9578214} & \cmark& \xmark & 3.42 & 5.03 & 3.28 & 3.91 &   4.56 & \textbf{3.54} & 3.25 & 3.78 & 6.00 & 5.20 & 6.55 & 5.92 &  4.83 &5.28 &6.04 & 5.38\\
 MNN~\cite{Valle2021MultiTaskHP}& \xmark & \xmark  & 3.34 & 4.69 & 3.48 & 3.83 &  3.98 & 4.61 & 2.39 & 3.66& -& -& - &- & - & -& -&-\\ 
 RankPose~\cite{rankpose}& \cmark & \xmark & \textbf{3.26} & 4.72 & 3.23 & 3.74 &      4.54 & 5.61 & 3.05 & 4.40 &     4.40& 4.42 & 5.08 & 4.63 & 5.81 & 5.91 & 7.39 & 6.37 \\
 \midrule
 \netname && \xmark & 3.27 & \textbf{4.58} & \textbf{2.98} & \underline{\textbf{3.61}} & 3.23 & 5.32 & 2.78 & 3.78 & \textbf{4.33} & \textbf{4.17} & \textbf{5.06} & \underline{\textbf{4.52}} & \textbf{4.66} & 5.29 & \textbf{6.03} & \textbf{5.32} \\
 \netname360 & &\cmark  & 3.73 & 5.52 & 3.53 & 4.26 & 3.37 & 3.87 & 2.93 & \underline{\textbf{3.39}} & 5.18 & 4.70 & 6.04 & 5.31 & \textbf{4.64} & \textbf{4.57} & \textbf{5.34} & \underline{\textbf{4.85}}\\
 \bottomrule
 
 {

}
\end{tabular}
\caption{Comparisons with the state-of-the-art methods on the
AFLW2000 and BIWI dataset. All models are trained on the 300W-LP dataset. Results from methods with positive $I$ are generated by our own tests. Methods with negative $I$ are not or only partially open-source. Their results are claims from authors. Methods with positive $R$ target the prediction of a wider range of rotation.}
\label{table1}
\end{table*}
\subsection{Implementation details}
We implement our proposed network using PyTorch~\cite{NEURIPS2019_9015}. As backbone, we choose ResNet50~\cite{7780459} to enable a fair comparison with other methods~\cite{Ruiz2018FineGrainedHP, rankpose, Cao_2021_WACV,8444061, 10.1007/978-3-030-90439-5_43}, that chose the same feature extractor. The backbone's weights are pretrained with the ImageNet~\cite{deng2009imagenet} dataset. For the final layers, we choose a single fully connected layer with 6 outputs. 
The network is trained for 80 epochs with a batch size of 80 using the Adam optimizer with a learning of $1e^{-4}$. To exploit full generalization potential, we also extensively augment our training data using Albumentations~\cite{info11020125} by applying random horizontal flipping, random scaling and cropping, random rotation up to [-45, +45] degrees, random occlusions, and further image color operations including random blur, random brightness contrast changes, and random RGB shifts.

\subsection{Comparison with state-of-the-art}
In this section, we conduct a comprehensive comparison with the state-of-the-art. We start with a cross-dataset evaluation to analyze our model's generalization capabilities, followed by an intra-dataset experiment and a detailed error analysis for further performance assessment. 

\subsubsection{Cross-dataset evaluation}
In our first experiment, we want to evaluate our approach against the state-of-the-art methods. To the end, we train two models. The first model (\textit{6DRepNet}) will strictly follow the common training convention by using the synthetic 300W-LP dataset for training and the two real-world datasets AFLW2000 and BIWI for testing. This will provide comparable information about our method's performance of directly regressing a diminished rotation matrix. For another model (\textit{6DRepNet360}) we change the training setup by replacing the 300W-LP dataset for training with our combined dataset (CMU + 300W-LP) for full rotation appearance training. The remaining training configuration will remain the same to place the focus on the impact of the enriched training data for the evaluation. We provide numerical results in Mean Absolute Error (MAE) of the Euler angles and in Mean Absolute Error of the rotation matrix vectors (MAEV). Most of the current state-of-the-art methods don't provide the MAEV for their results, so we retest these methods that have been open sourced and calculate the MAEV based on the converted rotation matrices for each test sample. 

Table \ref{table1} shows the results from the two model setups along with the results from other methods from the recent literature. For better interpretation, we added an extra column ($R$) to show which methods are trained to predict a larger range of rotations and which ones restrict their predictions to frontal poses. From the 15 listed methods, only two approached the exceeding of narrow angle range head pose estimation. \\
\\
\textbf{6DRepNet}\\
The table demonstrates that our model that was solely trained on the 300W-LP dataset outperforms all other methods on the AFLW2000 test dataset and surpasses the current top performer RankPose on AFLW2000 in Euler and vector errors. Besides the overall error rate, our model achieves top performing results for the pitch and roll error and equal results to the best reported yaw error. This indicates a very stable network learning, resulting in robust prediction properties. On the BIWI dataset, it achieves competitive results in respect to MAE and best results in respect to MAEV. The latter ought to be considered with caution, as there are no MAEV results reported for the MAE top performers.\\
\\
\textbf{6DRepNet360}\\
Our second model, 6DRepNet360, achieves very competitive results on AFLW2000 and even new state-of-the-art results on BIWI by surpassing WHENet-V by 3\%. Noticeably, this model only differs in its training data, where the added data aims to expand the predictable detection range of the yaw rotation. Yet, these samples include numerous stronger pitch rotations than 300W-LP (see Fig.~\ref{fig:dataset_distribution}). We argue that these samples benefit the model's performance for processing the challenging poses from the BIWI dataset, as the error for the pitch is reduced by 33\% compared to our the solely on 300W-LP trained model. Remarkably, WHENet is also trained for wide yaw predictions and is therefore most suitable to compare it with our 6DRepNet360. While WHENet is reported to perform even worse than its 300W-LP equivalent WHENet-V, our 6DRepNet360 model achieves over 20\% lower error rates on AFLW2000 and over a 10\% higher accuracy on BIWI. We believe that our choice of the 6D rotation matrix as rotation representation instead of WHENets Euler angle has a major impact on our superior results. In terms of rotation representation, TriNet is the most similar method to ours. But in contrast to our 6 parameter approach, they predict the entire 9 parameter rotation matrix and use an SVD to find an orthogonal-constrained solution. We argue that our more efficient approach leads to a higher reported accuracy.   

Fig~\ref{fig:qual_results} shows qualitative results from our 6DRepNet360 model. The first row illustrates prediction on test images from the AFLW2000 dataset with strong varieties of background, lightning, and camera angle. The second row shows test results with very strong head rotations from the CMU Panoptic test set that exceed the common [-99°,+99°] restrictions. In contrast to AFLW2000, it is captured in a laboratory environment with consistent lightning conditions and background. Nevertheless, 6DRepNet robustly predicts the head poses from varying camera angles. A very noteworthy example is the rightmost test image, as it presents a very challenging instance. While for frontal faces even stronger rotated poses provide meaningful features, visual cues are in this example mainly restricted to the head's shape. Yet, our model is able to predict reliable orientations even for these challenging kind of head poses.

\begin{table}[]
\setlength{\tabcolsep}{10pt}
\begin{tabularx}{\linewidth} 
{     l l  c   c  c   a }
\toprule
& & \multicolumn{4}{c}{\textbf{BIWI Euler}}\\
\midrule
 & I & Yaw & Pitch & Roll & MAE  \\
  \midrule
 HopeNet~\cite{Ruiz2018FineGrainedHP}&\cmark & 3.35 & 4.66 & 3.00 & 3.67 \\ 
 FSA-Net~\cite{Yang_2019_CVPR}& \cmark & 6.79 & 9.18 & 4.56 & 6.84 \\
 FDN~\cite{Zhang2020FDNFD}& \xmark &3.00 & 3.98&  2.88&  3.29\\
 MDFNet~\cite{9435939}&\xmark & 2.99 & 3.68 & 2.99 & 3.22 \\
 TriNet~\cite{Cao_2021_WACV} & \cmark&2.79 & 3.28 & 2.53 & 2.87\\
 DDD-Pose~\cite{10.1007/978-3-030-90439-5_43}&\xmark & 3.04 & \textbf{2.94} & 2.43 & 2.80 \\
  \midrule
\netname && \textbf{2.39} & 2.96 & \textbf{2.05} & \textbf{2.47} \\


\bottomrule
\end{tabularx}
\caption{Euler error comparisons with the state-of-the-art methods on the
70/30 BIWI dataset. Results from methods with positive $I$ are generated by our own tests. Methods with negative $I$ are not or only partially open-source. Their results are claims from authors}
\label{table2}
\end{table}

\begin{table}[]
\setlength{\tabcolsep}{10pt}
\begin{tabularx}{\linewidth} 
{    l  c  c    c  c   a }
\toprule
& & \multicolumn{4}{c}{\textbf{BIWI Vector}}\\
\midrule
  & I &  Left & Down & Front & MAEV\\
  \midrule
  FSA-Net~\cite{Yang_2019_CVPR} & \cmark & 9.09 & 10.19 & 11.26 & 10.18\\
 HopeNet~\cite{Ruiz2018FineGrainedHP} & \cmark&  5.55 & 5.64 & 5.78 & 5.66 \\ 
 TriNet~\cite{Cao_2021_WACV} & \cmark & 4.12 & 4.47 & 4.24 & 4.28\\
  \midrule
\netname & &  \textbf{3.39} & \textbf{3.27} & \textbf{3.89} & \textbf{3.52}\\

\bottomrule
\end{tabularx}
\caption{Vector error comparisons with the state-of-the-art methods on the
70/30 BIWI dataset. Results from methods with positive $I$ are generated by our own tests. Methods with negative $I$ are not or only partially open-source. Their results are claims from authors }
\label{table3}
\end{table}

\begin{table}[]
\setlength{\tabcolsep}{10pt}
\begin{tabularx}{\linewidth} 
{    l c c    c  c   a }
\toprule
&& \multicolumn{4}{c}{\textbf{CMU Panoptic}}\\
\midrule
  & I&  Yaw & Pitch & Roll & MAE\\
  \midrule
 Viet et al.~\cite{Viet2021SimultaneousFD}& \xmark &9.55 & 11.29& 8.32 & 9.72\\ 
 WHENet~\cite{Zhou2020WHENetRF} & \xmark & 8.51 & 7.67 & 6.78 & 7.66 \\
 \midrule
\netname360&  &  \textbf{2.08} & \textbf{3.16} & \textbf{2.75} & \textbf{2.66}\\


\bottomrule
\end{tabularx}
\caption{Model performance on the CMU Panoptic + 300W-LP combined dataset. 70\% of the dataset is used for training and the remaining 30\% for testing. Results from methods with positive $I$ are generated by our own tests. Methods with negative $I$ are not or only partially open-source. Their results are claims from authors.}
\label{table6}
\end{table}
\begin{figure*}[]
    \centering
    \begin{subfigure}{.33\linewidth}
        \includegraphics[width=\linewidth]{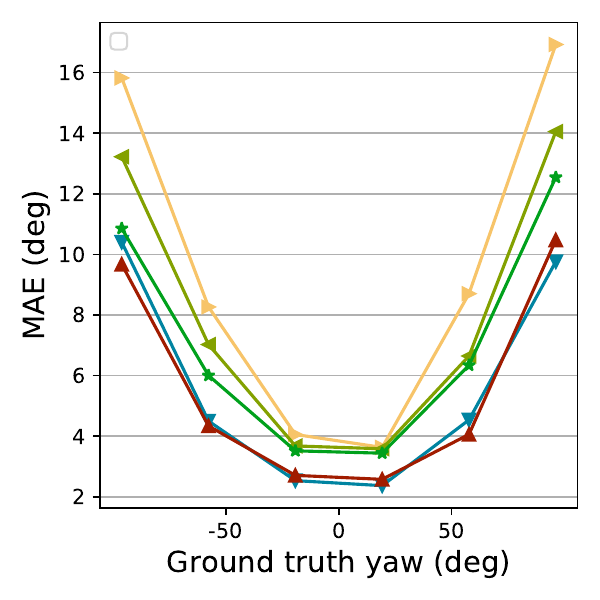}
    \end{subfigure}
    \begin{subfigure}{.33\linewidth}
        \includegraphics[width=\linewidth]{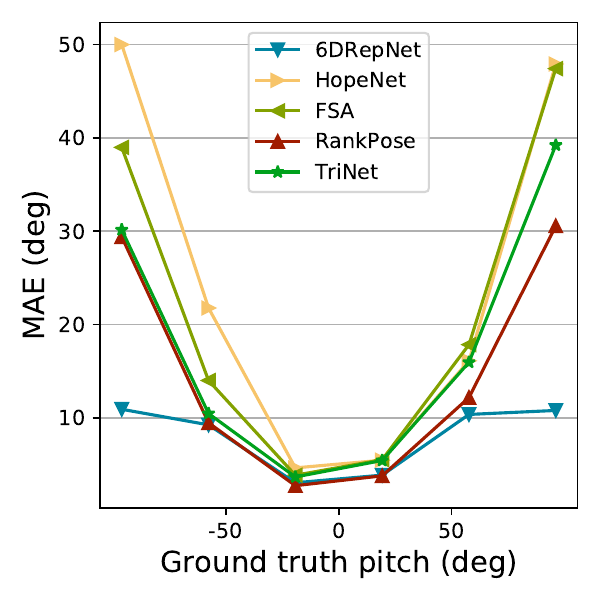}
    \end{subfigure}
    \begin{subfigure}{.33\linewidth}
        \includegraphics[width=\linewidth]{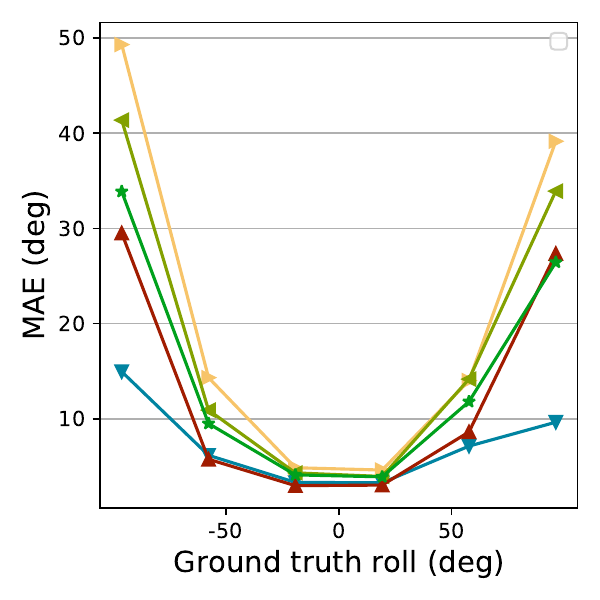}
    \end{subfigure}
    \caption{Error analysis for angle intervals on the AFLW2000 dataset.}
    \label{fig:error_analysis}
\end{figure*}
\subsubsection{Intra-dataset evaluation}
\label{sec:intra-dataset}
\textbf{BIWI}\\
In a second experiment, we follow the convention by FSA-Net~\cite{Yang_2019_CVPR} and randomly split the BIWI dataset in a ratio of 7:3 for training and testing, respectively. Table \ref{table2} and Table~\ref{table4} show our results compared with other state-of-the-art methods that followed the same testing strategy. We retested those models, that provide source code information, for an additionally MAEV error report. The remaining results are claims by the authors. It demonstrates that our method outperforms all other methods by a margin of more to 10\%. In terms of the individual rotation angles, our approach produces very consistent results by achieving the best results on yaw and roll, and equal results to the state-of-the-art DDD-Pose for the pitch angle. This supports the observed robustness in the cross-dataset evaluation and demonstrates, that achieving stable accurate results for all three angles does not only depend on the trained dataset, but rather on our proposed method itself. This is also reflected in Table~\ref{table4}, where our approach achieves the best overall MAEV results as well as for each single vector. \\
\\
\textbf{CMU Panoptic + 300W-LP}\\
In a final experiment, we evaluate our model in an intra-dataset test on our combined dataset that comprises the data from the CMU Panoptic and the 300W-LP dataset. To this end, we randomly split the dataset into 70\% training data and 30\% test data. To the best of our knowledge, Viet et al.~\cite{Viet2021SimultaneousFD} and WHENet are the only methods that published test results on CMU Panoptic. However, \cite{Viet2021SimultaneousFD}'s prediction pipeline additionally includes face detection and their test set comprises solely samples from CMU Panoptic. Therefore, the comparison ought to be considered with caution. More similar to our experimental approach, WHENet tests on a combination of CMU Panoptic and 300W-LP, but its size and composition are not specified. Thus, our results are mainly for future reference, and we will publish our test list to provide other methods with the capability of precise comparison. 

\subsubsection{Error Analysis}
To receive a more detailed impression of our model performance, we conduct an error analysis with four other state-of-the-art methods (HopeNet, FSA-Net, RankPose, TriNet) where we split up the errors on the AFLW2000 of range [-99°, 99°] into intervals of 33°. All models were solely trained on 300W-LP. The results are shown in Fig.~\ref{fig:error_analysis} where each Euler angle is illustrated in a separate graph. It gives insight that in general the prediction error for all methods increases with stronger rotations. It is conspicuous, though, that this error increase is much lower for 6DRepNet compared to all the other methods, especially for the pitch and roll. While Table~\ref{table1} shows that our model overall outperforms RankPose by 3\%, this detailed error analysis illustrates that our 6DRepNet achieves over 60\% smaller error rates for extreme pitch and roll rotations. This is yet another confirmation that our model does not only achieve state-of-that results, but at the same time provides very robust predictions even in extremely challenging test cases.  

\subsection{Ablation Study}
In the following, we will analyze how each of our model's remaining components impacts our reported results. This includes the backbone, that is responsible for the feature extraction, and our proposed loss function, which differs from other methods in the literature.
\subsubsection{Loss function} Most current methods use the Mean Squared Error (MSE)
\begin{equation}
    L_{MSE}=\frac{1}{N}\sum_{i=0}^{N}(y_{p}-y_{gt})^2
\end{equation}

for calculating the loss in the training procedure. We argue that the geodesic distance gives a better feedback about the distance between prediction and ground truth and, thus, is better suited to be used as a loss function. To prove this, we conduct another experiment where we repeat our previous tests, but this time we train our network with the MSE distance loss and with a combination of  MSE and the geodesic loss $L_g$ (see Eq~\ref{label:geodesicloss}). Table \ref{table3} shows these results compared to our models trained with geodesic distance loss. It states that the network with geodesic loss penalty performed significantly better than the one that used MSE and slightly better than the combination of MSE and $L_g$. 
\subsubsection{Backbone} In a final experiment, we analyze the impact of the chosen backbone on the results. Our results from table~\ref{table1} already proved the superiority of our 6D rotation matrix approach over other methods using the same backbone. Nevertheless, we want to evaluate the impact of the number of parameters on our results. In table~\ref{table:backbone} we compare our previous results with a model that was trained with the smaller ResNet18. It is remarkable, that our model that was trained on the 50\% smaller backbone ResNet18 still achieves better results on the AFLW2000 dataset than all other methods from Table~\ref{sec:experiments} except one. For the BIWI dataset, the accuracy compared to ResNet50 is reduced only by a very small margin. This confirms that our model's overall performance is predominantly accounted by our 6D rotation representation and hardly by the used backbone. Moreover, it shows that the commonly used ResNet50 is not necessary for achieving proper accuracy, as the more efficient ResNet18 reports similar performance. This becomes an important aspect, when the head pose estimation is used in settings with limited computational resources. 

\begin{table}[t]
\setlength{\tabcolsep}{3pt}

\begin{tabularx}{\linewidth} 
{    l  @{\hskip .2in}  c   c  c  a @{\hskip .22in} c c c a}
\toprule
&\multicolumn{4}{c}{\textbf{AFLW2000}} & \multicolumn{4}{c}{\textbf{BIWI}} \\
\midrule
& Yaw & Pitch &  Roll & MAE  & Yaw & Pitch &  Roll & MAE \\
  \midrule
$L_{MSE}$ & 3.38 & 4.89 & 3.33 & 3.87 &  3.19 & 6.52 & 2.81 & 4.17\\
$L_g$ +$L_{MSE}$ & 3.26 & 4.65 & 3.09 & 3.67 & 3.17 & 5.69 & 2.76 & 3.88 \\ 
$L_g$ & 3.27 & 4.58 & 2.98 & 3.61 & 3.23 & 5.32 & 2.78 & 3.78 \\

\bottomrule
\end{tabularx}
\caption{Analysis of the influence of different loss functions $L_{MSE}$ and geodesic loss $L_g$ on the MAE.}
\label{table4}
\end{table}
{
}

\begin{table}[]
\setlength{\tabcolsep}{4pt}

\begin{tabularx}{\linewidth} 
{    l  @{\hskip .2in}  c   c  c  a @{\hskip .2in}  c c c a}
\toprule
&\multicolumn{4}{c}{\textbf{AFLW2000}} & \multicolumn{4}{c}{\textbf{BIWI}} \\
\midrule
& Yaw & Pitch &  Roll & MAE  & Yaw & Pitch &  Roll & MAE \\
  \midrule
ResNet18 &  3.18 & 4.81 & 3.26 & 3.75 & 3.09 & 5.94 & 2.93 & 3.99 \\ 
ResNet50 & 3.27 &  4.58 & 2.98 & 3.61 & 3.23 & 5.32 & 2.78 & 3.78 \\

\bottomrule
\end{tabularx}
\caption{Comparison of the MAE between the different backbones.}
\label{table:backbone}
\end{table}

\begin{figure}[]
    \centering
    \includegraphics[width=\linewidth]{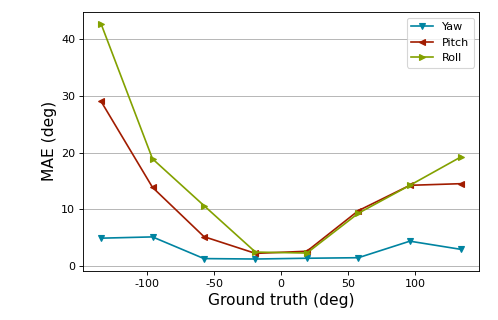}
    \caption{Euler angle error for the CMU Panoptic + 300W-LP test set.}
    \label{fig:cmu_error}
\end{figure}
\subsection{Limitations}
Our model achieves accurate and robust prediction for an extended range of rotation. This especially applies for the yaw angle, which encounters the strongest rotations in common application scenarios. However, the roll and pitch can also reach strong rotations, that are only marginally represented in our training data (see Fig.~\ref{fig:dataset_distribution}). This can lead to reduced robustness and accuracy in application scenarios with unusual camera angles and head poses. To analyze this, we degreewise calculated the error of our 6DRepNet360 model on the test set of our CMU Panoptic + 300W-LP 70/30 split from section~\ref{sec:intra-dataset}. The results are shown in Fig.~\ref{fig:cmu_error} and illustrate that the error rate for the yaw angle is consistently low, while the roll and pitch error rate increase with stronger rotations. This demonstrates that there is still a lack of training and also test data for this extended range of rotations. In our test set, only 3 samples exceed [-100,+100] degrees in roll and only 5 samples exceed [-100°,+100°] in pitch. In our experiments, we approached this limitation by performing image rotation augmentation that synthetically expands the roll and pitch range. Further, the CMU Panoptic dataset is taken in laboratory settings with similar background and lightning conditions. Additional data with stronger variation could therefore benefit the generalization performance as well.

\section{Conclusion}
In this paper, we tackle the major challenge of unconstrained full rotation head pose estimation that is a rarely explored research subject yet. First, we formulate a continuous 6D rotation matrix representation for an unambiguous and continuous appearance parameterization. This approach forms the basis for a stable and precise network training that we further optimize by introducing a geodesic distance based loss. With the use of the CMU Panoptic dataset, we accumulate a more comprehensive head pose dataset that exceeds the common public dataset in variety and size and allows us to create a model that is able to predict full head pose rotations. We evaluate our approach in multiple experiments that demonstrate that our 6D rotation representation achieves superior performance compared to the state-of-the-art and is able to efficiently learn the full range of head pose orientation. We complete our study with an ablation study to analyze the impact of the backbone and loss function on our results.


%

\ifCLASSOPTIONcompsoc
  \section*{Acknowledgments}
\else
  \section*{Acknowledgment}
\fi

This research was funded by the Federal Ministry of Education and Research of Germany (BMBF) project AutoKoWaT,~no.~13N16336 and by the German Research Foundation (DFG) project AL 638/15-1.

\ifCLASSOPTIONcaptionsoff
  \newpage
\fi



\bibliographystyle{IEEEtran}
\bibliography{bib}
%

%

%

\begin{IEEEbiography}[{\includegraphics[width=1in,height=1.25in,clip,keepaspectratio]{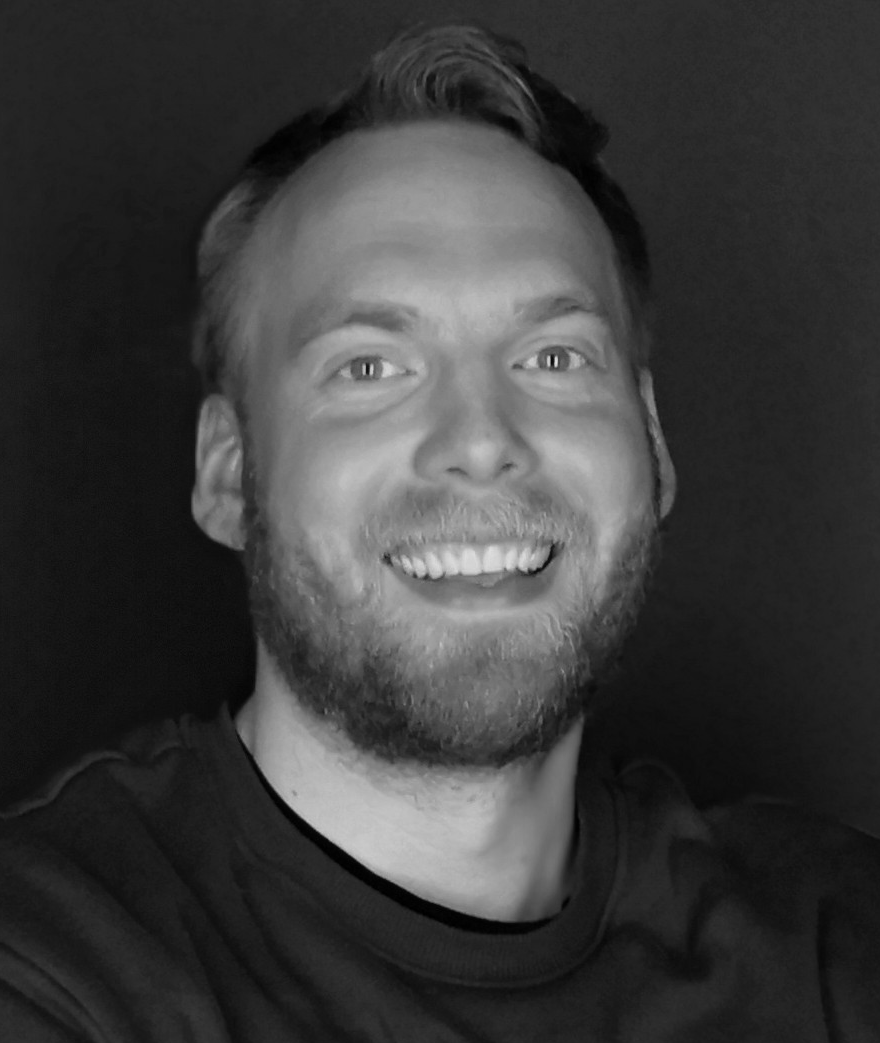}}]{Thorsten Hempel}
was born in Pinneberg, Schleswig-Holstein, Germany, in 1993. He received the B.S. degree in industrial engineering from the West Coast University of Applied Sciences, Heide, Germany, in 2017, and the M.S. degree in industrial engineering from the Otto-von-Guericke-University of Magdeburg, Germany, in 2019. His research interests include the field of computer vision, cognitive robotics and human–robot interaction. 
Since 2019 he has been a Research Assistant with the Neuro-Information Technology research group at the Otto-von-Guericke University Magdeburg.
\end{IEEEbiography}

\begin{IEEEbiography}[{\includegraphics[width=1in,height=1.25in,clip,keepaspectratio]{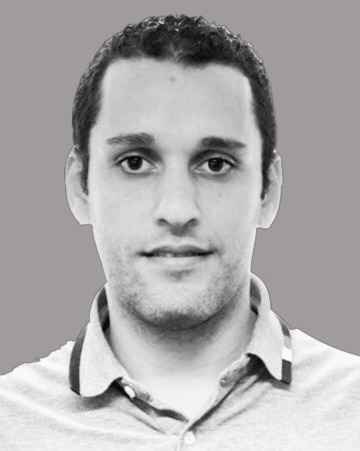}}]{Ahmed A. Abdelrahman} 
was born in Cairo, Egypt, in 1989. He received his B.Sc. and M.Sc degrees in electrical and computer engineering from MTC. He is currently pursuing his Ph.D. degree in electrical engineering at the Otto-von-Guericke University Magdeburg. His research interests include computer vision, deep learning, and human-- machine interaction.
Since 2021 he has been a Research Assistant with the Neuro-Information Technology research group at the Otto-von-Guericke University Magdeburg.
\end{IEEEbiography}


\begin{IEEEbiography}[{\includegraphics[width=1in,height=1.25in,clip,keepaspectratio]{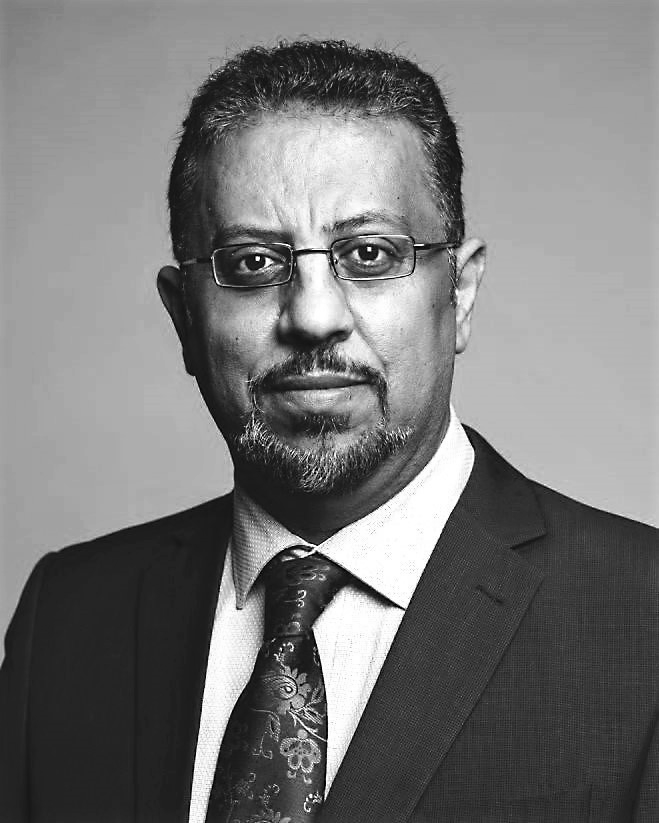}}]{Ayoub Al-Hamadi} 
received the Ph.D. degree in technical computer science, in 2001, and the Habilitation degree in artificial intelligence and the Venia Legendi degree in pattern recognition and image processing from Otto-von-Guericke University Magdeburg, Germany, in 2010. He is currently an Adjunct Professor and the Head of the Neuro-Information Technology Group, Otto-von-Guericke University Magdeburg. He is the author of more than 380 papers in peer-reviewed international journals, conferences, and books. His research interests include computer vision, pattern recognition, and image processing. See \href{http://www.nit.ovgu.de/}{http://www.nit.ovgu.de/} for more details.
\end{IEEEbiography}




\end{document}